\documentclass{article}

     \usepackage[final]{neurips_2019}

\usepackage[utf8]{inputenc} 
\usepackage[T1]{fontenc}    
\usepackage{hyperref}       
\usepackage{url}            
\usepackage{booktabs}       
\usepackage{amsfonts}       
\usepackage{nicefrac}       
\usepackage{microtype}      

\title{GDPR compliant collection of Therapist-Patient-Dialogues}

\author{%
Tobias Mayer$^{1}$, Neha Warikoo$^{1}$ , Oliver Grimm$^{2}$, Andreas Reif$^{2}$, Iryna Gurevych$^{1}$ \\ \\
   $^1$Ubiquitous Knowledge Processing Lab (UKP Lab) \\
   Department of Computer Science and Hessian Center for AI (hessian.AI) \\
   Technical University of Darmstadt \\
   \url{www.ukp.tu-darmstadt.de} \\ \\
   $^2$ Department of Psychiatry, Psychosomatic Medicine and Psychotherapy\\
   University Hospital Frankfurt - Goethe University\\
  \\
}

\begin{document}

\maketitle

\begin{abstract}
 
According to the Global Burden of Disease list provided by the World Health Organization (WHO), mental disorders are among the most debilitating disorders. To improve the diagnosis and the therapy effectiveness in recent years, researchers have tried to identify individual biomarkers. Gathering neurobiological data however, is costly and time-consuming. Another potential source of information, which is already part of the clinical routine, are therapist-patient dialogues. While there are some pioneering works investigating the role of language as predictors for various therapeutic parameters, for example patient-therapist alliance, there are no large-scale studies. A major obstacle to conduct these studies is the availability of sizeable datasets, which are needed to train machine learning models. While these conversations are part of the daily routine of clinicians, gathering them is usually hindered by various ethical (purpose of data usage), legal (data privacy) and technical (data formatting) limitations. Some of these limitations are particular to the domain of therapy dialogues, like the increased difficulty in anonymisation, or the transcription of the recordings. In this paper, we elaborate on the challenges we faced in starting our collection of therapist-patient dialogues in a psychiatry clinic under the General Data Privacy Regulation of the European Union with the goal to use the data for Natural Language Processing (NLP) research. We give an overview of each step in our procedure and point out the potential pitfalls to motivate further research in this field.





\end{abstract}

\section{Introduction}
Mental health problems are highly prevalent conditions that are rising constantly worldwide ~\cite{world2017depression}. A large part of it is manifestations in behaviour and language. Naturally, language is used as a means to diagnose and treat these diseases through therapy. In recent years, with the advancement of Artificial Intelligence (AI), more and more Machine Learning (ML) methods are applied on therapy or other mental health data to support the fight against these diseases ~\cite{MLsurvey,Bucci2019}. In particular, methods from Natural Language Processing (NLP) are used to analyse the linguistic behaviour ~\cite{Tanana2021HowDY,aafjes2020language,flemotomos2018language} with an increasing interest from both the NLP community and psychotherapy research. \\
However, progress in this field strongly depends on the availability of adequate data. In theory therapist-patient dialogues, as they are happening during therapy, seem like a fruitful data source. In practice, on the other hand, collecting, processing and especially sharing of clinical data is a complicated matter ~\cite{harrigian2021}. \\
Current NLP works on therapy session data are therefore conducted on small size datasets by ML standards, which are usually not available to the public. The data is usually in-house ~\cite{tsakalidis2021} or clinical trial data with restricted access~\cite{can2016sounds,flemotomos2018language,gibson2017attention}. Some use commercial ~\cite{Lee2019IdentifyingTC,Tanana2021HowDY} or other proprietary datasets ~\cite{rojas2018} as well. This data scarcity has led to work on other non-clinical mental health data, like mock ~\cite{primock2022} or online consultations ~\cite{sun-etal-2021-psyqa,perez-rosas-etal-2018-analyzing}, or social media posts ~\cite{harrigian2021,clpsych2019}. \\
While these offer interesting research and application scenarios, professional therapy session dialogues remain hard to access with great research potential. To encourage more research on psychotherapy dialogue datasets, we share our experience in starting our own data collection with the goal to analyse it with NLP methods. In this paper, we describe what was required to collect therapist-patient dialogues in a psychiatric clinic under the General Data Protection Regulation\footnote{\url{https://gdpr.eu/tag/gdpr/}} (GDPR) of the European Union and share them with a contractual partner university for NLP research. We highlight issues we encountered when applying for ethical approval, the complexities of data protection compared to other healthcare data, and finally the technical challenges which impede working with such data.

\section{Background}
We provide a short overview of this research project to contextualize our objectives properly. 
The overall goal of the planned research is to investigate how NLP methods can help in mental health research. To this end, we want to analyse therapist-patient dialogues. The involved parties consist of (a) the department of psychiatry and psychotherapy at the university hospital and (b) the computer science (CS) department at a different university. Both universities signed a cooperation contract for this project and are bound by the terms and conditions of the GDPR. 
The data is being collected at the university hospital, anonymized, and then shared with the CS department to conduct ML experiments.

\section{Data Collection Procedure}
In this section, we elaborate more on the challenges we faced while securing the approval for the data collection. In particular, we first talk about getting the approval of the ethics committee, second about our data protection concept, and third about the technical challenges.
\subsection{Ethics Committee}
As with social media health or other sensitive data, collecting and working with therapist-patient dialogues as such needs an ethical approval~\cite{benton2017}. In our case, this was a two step process. First, the study needed to be approved by the ethics committee of the medical department at the university hospital, where the data is collected. This set up is comparably to an Institutional Review Board (IRB), where a group of medical personnel evaluates the study layout, and other ethical criteria, like the purpose of the data usage and associated risk/benefit.
Furthermore, we needed a second approval from the ethics committee of the university where the data processing and analysis is planned. There, the committee consists of people with various backgrounds. Contrary to the first evaluation, the focus was on the application of AI/NLP methods to process clinical data, as AI systems are not perfect and their results need to be treated with care. Especially, when it can have implications for vulnerable people. 

\paragraph{(1) Data Collection}
The first submission was done by the partners at the university hospital. It comprised a study protocol, an information sheet for the patients accompanied by a consent form, and a form of technical and organisational measures (TOMs). The study protocol is a detailed description of various aspects of the study, e.g., who is the leading PI, which institutions and researchers are involved, how is it financed, or what is the scientific background etc. Furthermore, the study details are listed. In particular, which parameters of the patients are recorded. In our case, these are the Clinical Global Impression Scale (CGI), Global Assessment of Functioning (GAF), Depressive Symptom Index (DSI-SS), severity of the psycho-pathological findings, as well as the psycho-pharmacological medication. Additionally, size of the study population, in- and exclusion criteria, as well as how the participants are recruited, and how the study is conducted, have to be explained. Usually, an important part is also the risk-benefit assessment. In our case the participants do not have any additional strain and the treatment is independent of the study without any study specific intervention. Therefore, there are no noteworthy risks for the participants. \\
Before any of the above data can be collected, patients must be properly informed about the study and give their consent to participate. For this, the information sheet explains the purpose, duration and procedure of the study, states the individual benefit and potential risks for the participant (in our case both were none), ethical and legal (incl. insurance) basis, and the data protection concept. In particular, the latter states instructions on how the data can be withdrawn from the study, and the respective rights the participants still hold over their data, e.g., the right of disclosure, correctness, or deletion. Notably, this information sheet must be written in layman's terms, so it is comprehensible for everybody. Additionally, the phrasing should not leave space for interpretations by the participants. Concretely in our case, this meant that we should not mention that we analyse the data for the risk of suicidality, since then the participants might think that the therapist considers them suicidal. \\
By signing the informed consent form the participants gives allowance for the data to be used for the purpose stated in the information sheet and only this purpose. Any further research on this data requires another updated informed consent. However, asking initially for general consent allows future data usage for other research projects conducted within the hospital. \\
Another important part is the data management, in particular the data protection, which needs to be laid out in the study protocol and the information sheet as well. While these documents contain a high-level description of critical points and how they need to be taken care of during the study, a more elaborate data protection concept needs to be validated by the responsible data protection officer from the hospital. To be GDPR compliant, it requires a form of the technical and organisational measures which describe each single step of the data (from collecting over processing to deletion). This topic is further elaborated in the \textit{Data Protection} section.
\paragraph{(2) Data Processing with AI}
While in some cases the approval of only one of the ethics committees of the participating institutions is sufficient, it was recommended in our case to also ask for approval from the second responsible committee as well. This was due to the application of AI methods on data collected from vulnerable/sick people. In addition to the material from the first request for approval, a detailed description of the AI research purpose needed to be given. It was highlighted that no AI methods would influence the treatment of participants at any time and that an ex-post analysis is conducted to explore if current AI/NLP methods could find a connection between linguistic behavior and certain medical conditions, i.e., risk of suicidality. Furthermore it needed to be explained that the corresponding results and the algorithms from the study are not a potential medical device according to the Medical Device Regulation\footnote{\href{https://eur-lex.europa.eu/legal-content/EN/TXT/HTML/?uri=CELEX:32017R0745&from=IT}{REGULATION (EU) 2017/745}} (MDR) Article 2 of the European Union. According to the MDR, software used for diagnosis, prevention, monitoring, prediction, or prognosis count as medical devices. As the project is planned as an explorative study with no application or previously mentioned usage in an actual clinical environment, it does not fall under the MDR. However, follow-up studies or software applications which may use the data to predict similar parameters in a clinical environment need to be evaluated not only for ethical approval again, but may also fall under the MDR which comes with additional terms and conditions, that need to be addressed at the appropriate time. \\

\subsection{Data Protection}
As previously mentioned, data protection is an important part of getting the ethics approval. The ethical approval, however, does not mean that the data protection concept is approved. It merely confirms that the points which needs to be addressed are being considered by the study coordinators. For the final evaluation, the responsible data protection officer has to approve the data protection concept i.e. if the proposed measures are sufficient and GDPR compliant. In general, data usage under the GDPR is a complicated topic. For the sake of space and simplicity, we highlight only issues which we think are particular to the collection/processing of therapist-patient dialogues and the transfer between two research institutes. \\
The aforementioned TOMs contain a step-wise description of each station the data passes through. It describes processes, systems, and measures which are implemented to guarantee a secure data storage and usage without data breaches. The TOMs are an important document to evaluate compliance with the GDPR. Our dataset comprises of therapy dialogues which contain sensitive information in a non-trivial format. Therefore, the first noteworthy step is the pseudonymisation of the audio recordings from the therapy sessions. This comprises names and addresses of all the speakers. While there are software tools for automated pseudonymisation, most of them work only on textual data and are also prone to errors. To minimize the risk, the data needs to be pseudonymized with a human in the loop.
This step is similar when working with other medical data. 
Similar to social media data~\cite{benton2017}, where one has to take care that one does not leak identifiable information via links and connections, natural conversations might contain events which identify the patient. Especially in the case of forensic psychiatry, these types of unique events might occur more often. Therefore, to move from pseudonymized to anonmyized data, these events need to be removed from the data by manual intervention. Besides the content of the dialogues, other personal identifiable information must also be anonymized. According to the GDPR Article 4(1)\footnote{\url{https://gdpr-info.eu/art-4-gdpr/}}, voice is also considered as an identifiable physiological property. This prevents sharing of the raw speech data between the institutes. A solution to this is the transfer of technical audio features which preserve speaker privacy~\cite{NAUTSCH2019}. \\
Even with anonymized data, transferring the data to another lab comes with certain conditions. The base for the transfer is the cooperation contract between the two universities. However, it must be ensured that the data is safe at any time and access is restricted to the project members only, who also need to sign non-disclosure agreements. This procedure is detailed in the TOMs: Audio features, as well as the compiled clinical parameters are compressed and encrypted with an AES-256 bit key. Password and data are shared separately. The former via protected email, the latter via a data protection compliant sync- \& share-system. A two-stage firewall and access control together with storing the data on isolated and encrypted hard drives warrants data confidentiality. After the transfer, ML experiments are conducted on the protected in-house computation cluster of the CS-research group. After the experiments and analyses have been completed, the data must be deleted on the CS department side. However, the data at the hospital must be stored for 10 years for legal reasons.


\subsection{Technical Challenges}
Implementing the previously explained measures comes with a few technical challenges. One of the challenge is because of the nature of the data, i.e., natural language dialogues. This impacts multiple stages of processing the data. For a role-based analysis of the dialogue, like the risk assessment of suicidality, one wants to have the conversation separated into distinct speech-turns of the patient and the therapist which are processed as separate units. This speaker diarization can be either done manually or automatically. However, the manual separation does not scale with increasing amounts of data. Automatic systems, on the other side, are prone to errors and are usually offered as cloud services. Uploading speech data into a third-party cloud, however, is against the GDPR. A simple solution for the diarization problem is recording with the two microphones where each speaker has its own distinct audio channel. The trade-off for this method is a more complex setup in the clinical routine compared to one recording device used to record therapy sessions. \\
The second challenge is the quality of the recorded data. While in theory the recording setup (speaker distance to microphone, background noise, speaker volume, etc.) should be similar for each session, in practice this cannot be achieved. These fluctuations can be partially counter-balanced with post-processing the audio. This can include compression, frequency cleaning or limiting and noise suppression. \\
Another factor to consider is the amorphous character of these dialogues. Voices can be overlapping. Utterances are highly contextual, elliptical, and potentially ungrammatical. In our multicultural society, a variety of regional dialects and different accents are omnipresent. This complicates language processing, especially for speech processing, like automatic transcription systems. Given the goal is to combine acoustic features with text, the recorded sessions need to be transcribed. A variety of softwares are available for automated transcription. However, most of them are cloud services and show mediocre performance for clinical conversations ~\cite{primock2022}. Only few of them provide on-site execution. Similar to the pseudonymisation, manual transcription provides higher quality, but comes at a higher price in time and money with  human-in-the-loop approach. 

\section{Conclusion}
In this paper, we shared our experience with collecting therapist-patient dialogues in a psychiatric hospital under the GDPR. We highlighted the difficulties when applying for ethical approval, the specificities of the data protection when working and sharing this kind of data, and finally, the technical challenges the data poses. While there are multiple steps which we assigned to humans, some of them, like the diarization and transcription, can be in general automated at the cost of decreased data quality. The completeness of the pseudonymization and anonymisation, however, needs to be confirmed by a human rater. We hope that our work shows that interdisciplinary research collaborations in this domain under the GDPR are not impossible and motivate others to take the extra steps to do the same.




\bibliography{refs}
\bibliographystyle{abbrvnat}

\medskip

\small

\end{document}